\documentclass{article}
\usepackage{ijcai19}

\usepackage{times}
\usepackage{soul}
\usepackage{url}
\usepackage{hyperref}
\usepackage[utf8]{inputenc}
\usepackage[small]{caption}
\usepackage{graphicx}
\usepackage{amsmath}
\usepackage{booktabs}
\usepackage[algoruled]{algorithm2e}
\title{Towards White-box Benchmarks for Algorithm Control}

\author{
	André Biedenkapp\footnote{Contact Author}$^{,1}$\and
	H. Furkan Bozkurt$^{1}$\and
	Frank Hutter$^{1,2}$\And
	Marius Lindauer$^{1}$\\
	\affiliations
	$^1$ University of Freiburg ~~~~~~~ $^2$ Bosch Center for Artificial Intelligence\\
	\emails
	\{biedenka, bozkurth, fh, lindauer\}@cs.uni-freiburg.de
}

\usepackage{mathtools, nccmath}
\usepackage{latexsym}
\usepackage{amsmath,amssymb,amsfonts,dsfont}
\usepackage{booktabs}
\usepackage[usenames,dvipsnames]{xcolor, colortbl}
\definecolor{Gray}{gray}{0.875}
\usepackage{subfig}

\usepackage{url}
\usepackage[titletoc,title]{appendix}
\usepackage{ifthen}
\usepackage{lipsum}

\usepackage{pgfplots}
\usepackage{tikz}
\usetikzlibrary{shapes.geometric}
\usetikzlibrary{positioning,shapes,shadows,arrows,calc}
\usetikzlibrary{intersections}
\pgfplotsset{compat=1.14}
\pgfdeclarelayer{background}
\pgfdeclarelayer{foreground}
\pgfsetlayers{background,main,foreground}

\tikzstyle{activity}=[rectangle, draw=black, rounded corners, text centered, text width=6em, fill=white, drop shadow]
\tikzstyle{data}=[rectangle, draw=black, text centered, fill=black!10, text width=6em, drop shadow]
\tikzstyle{myarrow}=[->, thick]

\usepackage{xspace}
\usepackage{xstring}

\newcommand{\policy}[0]{\pi}

\newcommand{\insts}[0]{\mathcal{I}}
\newcommand{\instance}[0]{\mathit{i}}  

\sfcode`\.=1001
\sfcode`\?=1001
\sfcode`\!=1001
\sfcode`\:=1001
\newcommand{\autocase}[1]{\ifnum\ifhmode\spacefactor\else2000\fi>1000 \uppercase{#1}\else#1\fi}
\newcommand{\agent}[1][]{\autocase{c}ontroller#1\xspace} 

\newcommand{\target}[1][]{\autocase{a}lgorithm#1\xspace}
\newcommand{\atarget}[1][]{\autocase{a}n \target[#1]}

\newcommand{\param}[1][]{\autocase{h}yperparameter#1\xspace}

\newcommand{\task}[1][]{\autocase{i}nstance#1\xspace}
\newcommand{\atask}[1][]{\autocase{a}n \task[#1]}

\DeclareMathOperator*{\argmax}{arg\,max}
\newcommand{\citeN}[1]{\citeauthor{#1}~(\citeyear{#1})}

\let\oldcite=\cite
\renewcommand\cite[1]{\ifthenelse{\equal{#1}{NEEDED}}{[{\color{blue}citation~needed}]}{\oldcite{#1}}}

\newcommand{\states}{\mathcal{S}}
\newcommand{\actions}{\mathcal{A}}
\newcommand{\transitions}{\mathcal{T}}
\newcommand{\rewards}{\mathcal{R}}
\newcommand{\mdp}{\mathcal{M}}

\begin{document}

\maketitle

\begin{abstract}
    The performance of many algorithms in the fields of hard combinatorial problem solving, machine learning or AI in general
    depends on tuned \param configurations.
    Automated methods have been proposed to alleviate users from the tedious and error-prone task of manually searching for performance-optimized configurations across a set of problem instances.
    However there is still a lot of untapped potential through adjusting an algorithm's \param[s] online
    since different \param[s] are potentially optimal at different stages of the algorithm.
    We formulate the problem of adjusting an algorithm's \param[s] for a given instance on the fly as a contextual MDP,
    making reinforcement learning (RL) the prime candidate to solve the resulting \emph{algorithm control problem} in a data-driven way.
    Furthermore, inspired by applications of algorithm configuration, we introduce new white-box benchmarks suitable to study algorithm control.
    We show that on short sequences, algorithm configuration
    is a valid choice, but that with increasing sequence length a black-box view on the problem quickly becomes infeasible and RL performs better.
\end{abstract}

\section{Introduction}
    To achieve peak performance of an algorithm, it is often crucial to tune its \param[s].
    Manually searching for performance-optimizing \param configurations
    is a complex and error prone task. General algorithm configuration tools
    \cite{ansotegui-cp09a,hutter-lion11a,lopez-ibanez-orp16} have been proposed to 
    free users from the manual
    search for well-performing \param[s]. Such tools have been successfully applied to state-of-the-art solvers of various problem domains, such as mixed integer programming~\cite{hutter-cpaior10a},
    AI planning \cite{fawcett-icasp11a}, machine learning~\cite{snoek-nips12a}, or propositional
    satisfiability solving \cite{hutter-aij17a}.
    
    One drawback of algorithm configuration, however, is that it only yields a fixed configuration that is used during the entire solution process of the optimized
    algorithm. 
    It does not take into account that most algorithms used in machine learning, satisfiability solving (SAT), AI-planning,
    reinforcement learning or AI in general are iterative in nature.
    Thereby, these tools ignore the possible
    induced non-stationarity of the optimal target \param configuration.

    We propose a general framework to learn to control algorithms which we dub
    \emph{algorithm control}. We formulate the problem of learning dynamic algorithm control policies wrt its \param[s] as a contextual Markov decision process (MDP)
    and apply reinforcement learning to it.
    Prior work that considered online tuning of algorithms did not explicitly take problem \task[s]
    into account~\cite{battiti-11} and did not pose this problem as
    a reinforcement learning problem~\cite{adriaensen-ijcai16}. 
    To address these missing, but important components, we introduce
    three new white-box benchmarks suitable for algorithm control.
    On these benchmarks we show that, using reinforcement learning, we are able to successfully
    learn dynamic configurations across instance sets directly from data, yielding better performance than static configurations.
    
    Specifically, our contributions are as follows:
    \begin{enumerate}
        \item
            We describe controlling algorithm \param[s] as a contextual MDP,
            allowing for the notion of instances;
        \item
            We show that black-box algorithm configuration is a well-performing option for learning
            short policies;
        \item
            We demonstrate that, with increasing policy length, even in the homogeneous setting, traditional algorithm configuration becomes in-feasible;
        \item
            We propose three new white-box benchmarks that allow to study algorithm control
            across instances;
        \item We demonstrate that we can learn dynamic policies across a set of instances showing the robustness of applying RL for algorithm control.
    \end{enumerate}

\section{Related Work}
    Since algorithm configuration by itself struggles with heterogeneous instance sets (in which different configurations
    work best for different instances), it was 
    combined with algorithm selection~\cite{rice76a} to search for multiple well-performing
    configurations and select which of these to apply to
    new instances~\cite{xu-aaai10a,kadioglu-ecai10}.
    For each problem instance, this more general form of per-instance algorithm configuration still uses fixed configurations.
    However for different AI applications, dynamic configurations can be
    more powerful than static ones.
	A prominent example for \param[s] that need to be controlled over time is the learning rate in
	deep learning: a static learning rate can lead to sub-optimal training
	results and training times~\cite{moulines-neurips11}. To facilitate fast training
	and convergence, various learning rate schedules or adaptation schemes have been proposed
	\cite{schaul-icml13,Kingma-iclr14,singh-ieee15,daniel-aaai16,loshchilov-iclr17a}. Most of these methods, however,
	are not data-driven.
	
    In the context of evolutionary algorithms, various online hyperparameter adaptation methods have been
    proposed~\cite{karafotias-ec15,doerr-arxi18}.
    These methods, however, are often tailored to one individual problem or rely on heuristics.
    These adaptation methods are only rarely learned in a data-driven fashion~\cite{sakurai-sitis2010}.
    
    Reactive search~\cite{battiti-book08} uses handcrafted heuristics to adapt an algorithms
    parameters online. To adapt such heuristics to the task at hand, hyper-reactive 
    search~\cite{ansotegui-aaai17} parameterizes these heuristic and applies per-instance algorithm configuration.

    The work we present here can be seen as orthogonal to work presented under the heading of
    learning to learn~\cite{andrychowicz-neurips16,li-arxiv16,chen-arxiv16}. Both lines of work intend to learn optimal instantiations of algorithms
    during the execution of said algorithm. The goal of learning to learn, however, is to learn an
    update rule in the problem space directly whereas the goal of
    algorithm control is to indirectly influence the update by adjusting the \param[s] used
    for that update. 
    By exploiting existing manually-derived algorithms and only controlling their hyperparameters well, algorithm control may be far more sample efficient and generalize much better than directly learning algorithms entirely from data.  

\section{Algorithm Control}
    In this section we show how algorithm control (i.e., algorithm configuration per time-step) can be
    formulated as a sequential decision making process.
	Using this process, we can learn a policy to configure \atarget's \param[s] on the fly, using reinforcement learning (RL).
%
	\subsection{Learning to Control Algorithms}
	We begin by formulating \emph{algorithm control} as a Markov Decision Process (MDP)
	$\mdp\coloneqq(\states,\actions,\transitions,\rewards)$. An MDP is a 4-tuple, consisting of a
	state space $\states$, an action space $\actions$, a transition function
	$\transitions$ and a reward function $\rewards$.
	
    		\paragraph{\textbf{State Space}}
    		
    		At each time-step $t$, in order to make informed
    		choices about the \param values to choose, the \agent needs to be informed about the
    		internal state $s_t$ of the \target{} being controlled. 
    		Many \target[s] collect various statistics that are available at
    		each time-step. For example, a SAT solver might track how many clauses are satisfied
    		at the current time-step. Such statistics are suitable to inform the \agent about the
    		current behaviour of the \target.
    		
            \paragraph{\textbf{Action Space}}
            
            Given a state $s_t$, the \agent has to decide how to change the value $v\in\actions_h$
            of a \param $h$ or directly assign a value to that \param,
            out of a range of valid choices. This gives rise to the overall action space
            $\actions = \actions_{h_1}\times\actions_{h_2}\times\ldots\times\actions_{h_n}$ for $n$ \param[s] of the algorithm at hand.
            
            \paragraph{\textbf{Transition Function}} 
            
            The transition function describes the dynamics of the system at hand. For example,
            the probability of reaching state $s_{t+1}$ after applying action $a_t$ in state $s_t$ 
            can be expressed as $p(s_{t+1}|a_t, s_t)$.
            For simple algorithms and a small instance space, it might be possible to derive the
            transition
            function directly from the source code of the \target. However, we assume that the
            transition function cannot not be explicitly modelled for interesting algorithms.
            Even if the dynamics are not modelled, RL can be used to learn an optimizing policy directly
            from observed transitions and rewards.
            
            \paragraph{\textbf{Reward Function}} 
            
            In order for the \agent to learn which actions
            are better suited for a given state, the \agent receives a reward signal
            $\rewards_i(s_t,a_t)\to\mathbb{R}$.
            On many RL domains the reward is sparse, i.e.,
            only very few state-action pairs result in an immediate reward signal. 
            If an algorithm already estimates the distance to some goal state well,
            such statistics might be suitable candidates for the
            reward signal, with the added benefit that such a reward signal is dense.
            
            \paragraph{\textbf{Learning policies}}
            
            Given the MDP $\mdp$ the goal of the \agent is to search for a policy
            $\policy^*$ such that
            \begin{align}
                 \policy^*(s)&\in
                        \argmax_{a\in\actions} \rewards(s,a)+\mathcal{Q}_{\policy^*}(s,a)\label{eq:opt_pol}\\
                 \mathcal{Q}_{\policy}(s,a)&=
                        \mathbb{E}_{\policy}\left[\sum_{k=0}^\infty\gamma^k r_{t+k+1}| s_t=s, a_t=a\right]\label{eq:actionstatefun}
            \end{align}
            where $\mathcal{Q}_{\policy}$ is the \emph{action-value function}, giving the
            expected discounted future reward, starting from state $s$, applying action $a$ and following
            policy $\policy$ with discounting-factor $\gamma$.
	
	\subsection{Learning to Control across Instances}
	    Algorithms are most often tasked with solving varied problem \task[s] from the same, or similar domains.
	    Searching for well performing \param settings on only one \task might lead to a strong performance on that \task but might not generalize to
	    new \task[s]. In order to facilitate generalization of algorithm control, we explicitly
	    take problem \task[s] into account. 
	    The formulation of algorithm control given above does not take instances into account, treating the problem
	    of finding well performing \param[s] as independent of the problem \task.
	    
	    To allow for algorithm control across instances, we formulate the problem as a
	    contextual Markov Decision Process
    	$\mdp_\instance\coloneqq(\states,\actions,\transitions_\instance,\rewards_\instance)$,
    	for a given \task $\instance\in\insts$.
    	This notion of context induces multiple
    	MDPs with shared action and state spaces, but with different transition and reward functions.
    	In the following, we describe how the context influences the parts of the MDP.
            
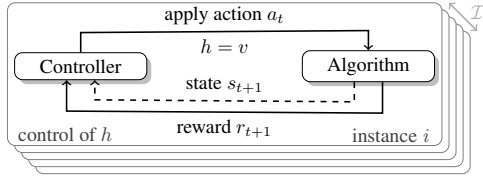
\begin{figure}[tbp]
        		\centering
        		\scalebox{.75}{
        			\begin{tikzpicture}[node distance=2.1cm]
        			
        			\node (Agent) [activity] {Controller};
        			
        			\node (Algo) [activity, right of=Agent, xshift=3cm] {Algorithm};
        			
        			\begin{pgfonlayer}{background}
        			\path (Agent -| Agent.west)+(-0.12,1.125) node (resUL) {};
        			\path (Algo.east |- Algo.south)+(0.125,-1.125) node (resBR) {};
        			
        			\path [rounded corners, draw=black!50, fill=white] ($(resUL)+(0.5, -0.5)$) rectangle ($(resBR)+(0.5, -0.5)$);
        			\path [rounded corners, draw=black!50, fill=white] ($(resUL)+(0.375, -0.375)$) rectangle ($(resBR)+(0.375, -0.375)$);
        			\path [rounded corners, draw=black!50, fill=white] ($(resUL)+(0.25, -0.25)$) rectangle ($(resBR)+(0.25, -0.25)$);
        			\path [rounded corners, draw=black!50, fill=white] ($(resUL)+(0.125, -0.125)$) rectangle ($(resBR)+(0.125, -0.125)$);
        			
        			\path [rounded corners, draw=black!50, fill=white] (resUL) rectangle (resBR);
        			\path (resBR)+(-.9,0.175) node [text=black!75] {\task $\instance$};
        			\path (resUL.east |- resBR.north)+(+.9,0.075) node [text=black!75] {control of $h$};
        			
        			\end{pgfonlayer}
        			
        			
        			\draw[myarrow] (Agent.north) -- ($(Agent.north)+(0.0,+0.35)$) -- ($(Algo.north)+(0.0,+0.35)$) node [above,pos=0.5] {apply action $a_t$} node [below,pos=0.5] {$h=v$} -- (Algo.north);
        			\draw[myarrow, dashed] ($(Algo.south)+(-0.25, 0)$) -- ($(Algo.south)+(-0.25, -0.35)$) -- ($(Agent.south)+(0.25, -0.35)$) node [above,pos=0.5] {state $s_{t+1}$} -- ($(Agent.south)+(0.25, 0)$);
        			\draw[myarrow] ($(Algo.south)+(0.25, 0)$) -- ($(Algo.south)+(0.25, -0.55)$) -- ($(Agent.south)+(-0.25, -0.55)$) node [below,pos=0.5] {reward $r_{t+1}$} -- ($(Agent.south)+(-0.25, 0)$);
        			
        			\draw[<->, thick, draw=black!32.5] (resBR.east |- resUL.center) -- ($(resBR.east |- resUL.center)+(0.5, -0.5)$);
        			\path (resBR.east |- resUL.center)+(.5, -0.175) node [text=black!32.5] {$\insts$};
        			\path (resUL.west |- resUL.center)+(-.5, -0.175) node [text=black!0] {$\insts$}; 
        			
        			\end{tikzpicture}
        	}
        		\caption{Control of \param $h$ of \atarget on a given contextual \task
        		    $\instance\in\insts$, at time-step $t\in T$.
        			Until \atask is solved or a maximum budget reached, the \agent decides which value $v$ to apply to \param $h$ based on the internal state $s_t$ of the \target, on the given \task $\instance$.}
        		\label{fig:pc}%
        	\end{figure}

            \paragraph{\textbf{Context}} 
            The \agent['s] goal is to learn a policy that can be applied to various problem \task[s] $\instance$ out of a set of \task[s] $\insts$. 
            We treat the \task at hand as context to the MDP.
	%
        	Figure~\ref{fig:pc} outlines the interaction between \agent and \target in that setting.
        	Given \atask $\instance$, at time-step $t$, the \agent applies action $a_t$ to the \target,
        	i.e., setting \param $h$ to value $v$.
        	Given this input, the \target advances to state $s_{t+1}$ producing a reward signal $r_{t+1}$, based on which
        	the \agent will make its next decision. The \task stays fixed during the \target run.

    		\paragraph{State and Action spaces}

            The space of possible states
    		does not change when switching between \task[s] from the same set, and is shared
    		between all MDPs induced by the context. Thus we consider the same state features.
    		To enrich the state space, we could also add \task-specific information, so-called instance features such as problem size,
    		which could be useful in particular for heterogeneous instance sets~\cite{leyton-brown-acm09a,hoos-lion12a}.

            %

            Similar to the state space, the action space stays fixed for all MDPs induced by the context.
            The action space solely depends on the \target at hand and is thus shared across all MDPs
            of the same context.
            
            \paragraph{\textbf{Transition Function}}

            Contrary to the state and action space, the transition function is influenced by the
            choice of the \task. For example, a search algorithm might be faced with completely
            different search spaces where applying an action could lead to different kind of states.
            %
            
            
            \paragraph{\textbf{Reward Function}} 

            As the transition function depends on the \task at hand, so does the reward function.
            Depending on the \task, transitions beneficial for the \agent on one \task might become unfavorable or might punish the agent on another \task. 
            
            It is possible to choose a proxy reward function that is completely independent of the
            context, i.e., a negative reward for every step taken. This would incentivize the \agent to learn
            a policy to quickly solve \atask which would be interesting if the real objective is to minimize runtime. 
            However, a \agent using such a reward would potentially take
            very long to learn a meaningful policy as the reward would not help it to easily distinguish between
            two observed states.
            
            \paragraph{\textbf{Learning policies across instances}}
            
            Given the MPD and a set of \task[s] $\insts$ the goal of the \agent is to find a policy~$\policy_{\instance\sim\insts}^*$ such that
            \begin{align}
                 \policy_{\instance\sim\insts}^*(s)&\in
                        \argmax_{a\in\actions} \rewards_{\instance\sim\insts}(s,a)+\mathcal{Q}_{\policy_{\instance\sim\insts}^*}(s,a)\label{eq:opt_pol_inst}\\
                 \medmath{\mathcal{Q}_{\policy_{\instance\sim\insts}}(s,a)}&=
                        \medmath{\mathbb{E}_{\policy_{\instance\sim\insts}}\left[\sum_{k=0}^\infty\gamma^k r_{\instance, t+k+1}| s_t=s, a_t=a, \instance\sim\insts \right]}\label{eq:actionstatefun_inst}
            \end{align}
            where $\mathcal{Q}_{\policy_{\instance\sim\insts}}$ is the \emph{action-value function}, giving the
            expected discounted future reward, starting from $s$, applying action $a$, following
            policy $\policy_{\instance\sim\insts}$ on \task $\instance$ with discounting-rate $\gamma$.
            
            \paragraph{Relation to Algorithm Configuration and Selection}
            This formulation of algorithm control allows to recover algorithm configuration (AC) as a special case: in AC, the optimal policy would simply always return the same action, for each state and instance. Further, this formulation also allows to recover per-instance algorithm configuration (PIAC) as a special case: in PIAC, the policy would always return the same action for all states, but potentially different actions across different instances. Finally, algorithm selection (AS) is a special case of PIAC with a 1-dimensional categorical action space that merely chooses out of a finite set of algorithms.

\section{Benchmarks}
    To study the algorithm control setting we 
    use two benchmarks already proposed by \citeN{adriaensen-ijcai16} and introduce three new benchmarks. Our proposed benchmarks increase the complexity of the
    optimal policy by either increasing the action space and policy length or including
    instances.
    
    
    
    \paragraph{Counting} The first benchmark introduced by \citeN{adriaensen-ijcai16} requires an agent to learn a monotonically increasing sequence.
    The agent only receives a reward if the chosen action has been selected at the corresponding
    time-step. This requires the agent to learn to count, where the size of the action
    space is equal to the sequence length. In the original setting of \citeN{adriaensen-ijcai16},
    agents need to learn to count to five, with the optimal policy resulting in a reward of five.
    The state is simply given by the history of the actions chosen so far.

    \paragraph{Fuzzy}
    The second benchmark introduced by \citeN{adriaensen-ijcai16} only features two actions.
    Action $1$ returns
    a fuzzy reward signal drawn from $\mathcal{N}(1, 2)$, whereas playing action $0$
    terminates the sequence prematurely. The maximum
    sequence length used in \citeN{adriaensen-ijcai16} is $20$ with an expected reward of the optimal
    policy also being $20$.
    Similar to the previous benchmark, \emph{Fuzzy} does not include any state representation other
    than a history over the actions.
    
    \paragraph{Luby}
    Similar to the already
    presented benchmarks, the newly proposed \emph{Luby} (see Benchmark Outline~\ref{bench:luby}) 
    does not model instances explicitly.
    However, it increases the complexity of learning a sequence compared
    to the benchmarks by \citeN{adriaensen-ijcai16}. An agent is required to learn the values in
    a Luby sequence~\cite{luby-ipl93}, which is, for example, used for restarting SAT solvers. The sequence is $1, 1, 2, 1, 1, 2, 4, 1, 1, 2, 1, 1, 2, 4, 8, ...$;
     formally, the $t$-th value in the sequence can be computed as:
    \begin{equation}\label{eq:luby}
        l_t = \left\{
	        \begin{array}{ll}
    	        2^{k-1}&  \mathrm{if }\,t = 2^k - 1,\\
    	        l_{t - 2^{k - 1} + 1}& \mathrm{if }\,2^{k-1}\leq t < 2^k - 1.
	        \end{array}
        \right.
    \end{equation}
    This gives rise to an action space for sequences of length $T$ with $\actions\coloneqq\left\{0,1, \ldots, \lfloor\log _{2}T\rfloor\right\}$ for all time-steps $t \leq T$, with the action values giving the exponents used in the Luby
    sequence.
    For such a sequence, an agent can benefit from state information about the sequence, such as
    the length of the sequence. For example, imagine an agent has to learn the Luby sequence for
    length $T=16$. Before time-step $8$ the action
    value $3$ would never have to be be played.
    For a real algorithm to be controlled,
    such a temporal feature could
    be encoded by the iteration number directly or some other measure of progress.
    The state an agent can observe therefore consists of such a time feature and a small history
    over the five last selected actions.
    {\SetAlCapNameFnt{\small}
    \SetAlCapFnt{\small}
    \SetAlCapSty{}
    \SetAlgorithmName{Benchmark Outline}{bench}{List of Benchmarks}
    \SetKw{kwActs}{Actions:}
    \SetKw{kwStat}{States:}
    \begin{algorithm}[tbp]
        \caption{Luby}
        \label{bench:luby}
        \kwActs{$a_t\in\left\{0,1, \ldots, \lfloor\log _{2}T\rfloor\right\}$ for all $0\leq t\leq T$}\;
        \kwStat{$s_t\in \left\{t, \mathrm{Hist}(a_{t-4}, a_{t-3}, \ldots, a_t)\right\}$}\;
        \For{$t\in\left\{0,1, \ldots, T\right\}$}{
            \uIf{$a_t==\mathrm{luby}(t)$}{
            reward$_t = 1$\;
            }
            \Else{reward$_t = -1$\;}
        }
    \end{algorithm}}
    
    \paragraph{Sigmoid}
    Benchmark Sigmoid (see Benchmark Outline~\ref{bench:sig})  allows to study algorithm control across instances. 
    Policies depend on the sampled instance $i$, which is
    described by a sigmoid 
    $sig(t; s_i,p_i)= \frac{1}{1 + e^{-s_i\cdot(t - p_i)}}$ that can be characterized through its
    inflection point $p_i$ and scaling factor $s_i$.
    The state is constructed using a time feature, as well as the instance information
    $s_i$ and $p_i$.
    
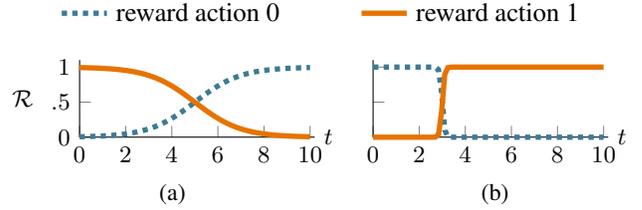
\begin{figure}[tbp]
        \centering
	    \pgfmathdeclarefunction{sig}{2}{%
	        \pgfmathparse{1/(1 + exp(-1*#1*(x-#2)))}}%
		\ref{graph1} reward action 0\qquad\quad\ref{graph2} reward action 1
	    {
	    \subfloat[]{
    		\centering
			\begin{tikzpicture}
			    \begin{axis}[
			            scale=0.3,
                        no markers,
                        domain=0:10,
                        samples=75,
                        axis lines*=left,
                        ymin=0.001,
                        ymax=1.1,
                        xmax=10.2,
                        xlabel=$t$,
                        ylabel=$\rewards$,
                        ylabel style={rotate=-90},
                        every axis x label/.style={at=(current axis.right of origin),anchor=west},
                        every x tick label/.append style={font=\small},
                        every y tick label/.append style={font=\small},
                        height=5cm,
                        width=12cm,
                        label style={font=\small},
                        xtick={0,2,4,6,8,10},
                        xticklabels={$0$,$2$, $4$,$6$,$8$, $10$},
                        yticklabels={$0$, $.5$, $1$},
                        ytick={0, .5, 1},
                        enlargelimits=false,
                        clip=false,
                ]
			        \addplot [line width=2pt,cyan!50!black,dotted] {sig(1,5)};
			        \addplot [line width=2pt,orange!90!black] {1-sig(1,5)};
		        \end{axis}
			\end{tikzpicture}\label{fig:sig_A}
		}%
	    \subfloat[]{
    		\centering
			\begin{tikzpicture}
			    \begin{axis}[
			            scale=0.3,
                        no markers,
                        domain=0:10,
                        samples=75,
                        axis lines*=left,
                        ymin=0.001,
                        ymax=1.1,
                        xmax=10.2,
                        xlabel=$t$,
                        every axis x label/.style={at=(current axis.right of origin),anchor=west},
                        every x tick label/.append style={font=\small},
                        every y tick label/.append style={font=\small},
                        height=5cm,
                        width=12cm,
                        label style={font=\small},
                        xtick={0,2,4,6,8,10},
                        xticklabels={$0$,$2$, $4$,$6$,$8$, $10$},
                        ytick={0, .5, 1},
                        yticklabels={ , , },
                        enlargelimits=false,
                        clip=false,
                ]
			        \addplot [line width=2pt,cyan!50!black,dotted] {1-sig(20,3)}; \label{graph1}
			        \addplot [line width=2pt,orange!90!black] {sig(20,3)}; \label{graph2}
		        \end{axis}
			\end{tikzpicture}\label{fig:sig_B}
		}}
		\caption{Example rewards for Benchmark~\protect\ref{bench:sig} with $T=10$ on
		both instances, $p_{(a)}=5, s_{(a)}=1$ on instance \protect\subref{fig:sig_A} and $p_{(b)}=3, s_{(b)}=20$
		on instance \protect\subref{fig:sig_B}.
		The solid line
		shows the received reward when playing action $1$ and the dashed line gives the reward
		for action $0$. On the $x$-axis are the time steps and on the $y$-axis is the reward.
		On instance \protect\subref{fig:sig_A} it is preferable to select action $1$ for the
		first halve of the sequence whereas on instance \protect\subref{fig:sig_B} it
		is better to start with action $0$.}
		\label{fig:sigmoids}%
	\end{figure}

    At each
    time-step an agent has to decide between two actions. The received reward when playing action
    $1$ is given by the function value of the sigmoid $sig(t; s_i, p_i)$ at time-step $t$ and $1 - sig(t; s_i, p_i)$
    otherwise. 
	The scaling of the sigmoid function is sampled uniformly at random in the interval
	$(-100, 100)$. The sign of the scaling factor determines if an optimal policy on the instance
	should begin by selecting action $0$ or $1$.
	The inflection point is distributed according to $\mathcal{N}(T/2, T/4)$ and determines how often an
    action has to be repeated before switching to the other action. Figure~\ref{fig:sigmoids}
    depicts rewards for two example instances. 
    The sigmoid in Figure~\ref{fig:sig_A}
    is unshifted and unscaled, leading to an optimal policy of playing action $1$ for the first
    half of the sequence and $0$ for the rest of the sequence. In Figure~\ref{fig:sig_B} the
    sigmoid is shifted to the left such that the inflection point is at $t=3$ and scaled by factor
    $20$. The optimal policy in this case is to play action $0$ for the first three steps and $1$
    for the rest of the sequence.
    
    {\SetAlCapNameFnt{\small}
    \SetAlCapFnt{\small}
    \SetAlCapSty{}
    \SetAlgorithmName{Benchmark Outline}{bench}{List of Benchmarks}
    \SetKw{kwActs}{Actions:}
    \SetKw{kwStat}{States:}
    \SetKwProg{Fn}{Function}{:}{}
    \begin{algorithm}[tbp]
        \caption{Sigmoid}
        \label{bench:sig}
        $s_i\sim\mathcal{U}(-100, 100)$\;
        $p_i\sim \mathcal{N}(T/2, T/4)$\;
        \kwActs{$a_t\in\left\{0,1\right\}$ for all $0\leq t\leq T$}\;
        \kwStat{$s_t\in\left\{t; s_i, p_i\right\}$}\;
        \For{$t\in\left\{0,1, \ldots, T\right\}$}{
            reward$_t = 1 - sig(t; s_i, p_i)$\;
            \If{$a_t==1$}{
                reward$_t = sig(t; s_i, p_i)$\;
            }
        }
    \end{algorithm}}
    \paragraph{SigmoidMVA}
        Benchmark SigmoidMVA (see Benchmark Outline~\ref{bench:sigmva}) further increases the complexity of learning across
        instances by translating the setting of \emph{Sigmoid} into a multi-valued action
        setting. An agent not only has to learn a simple policy switching between
        two actions but to learn to follow the shape of the sigmoid function used to
        compute the reward. The available actions an agent can choose from at each time-step
        are $a_t\in\left\{\frac{0}{L},\frac{1}{L},\ldots, \frac{L}{L}\right\}$. 
        Note that, depending on the granularity of the discretization
        (determined by $L$)
        the agent can follow the sigmoid more or less closely (thereby directly affecting its reward).
        
    {\SetAlCapNameFnt{\small}
    \SetAlCapFnt{\small}
    \SetAlCapSty{}
    \SetAlgorithmName{Benchmark Outline}{bench}{List of Benchmarks}
    \SetKw{kwActs}{Actions:}
    \SetKw{kwStat}{States:}
    \SetKwProg{Fn}{Function}{:}{}
    \begin{algorithm}[t]
        \caption{SigmoidMVA}
        \label{bench:sigmva}
        $s_i\sim\mathcal{U}(-100, 100)$\;
        $p_i\sim \mathcal{N}(T/2, T/4)$\;
        \kwActs{$a_t\in\left\{\frac{0}{L},\frac{1}{L},\ldots, \frac{L}{L}\right\}$ for all $0\leq t\leq T$}\;
        \kwStat{$s_t\in\left\{t, s_i, p_i\right\}$}\;
        \For{$t\in\left\{0,1, \ldots, T\right\}$}{
            reward$_t = 1 - abs(sig(t, s_i, p_i) - a_t)$\;
        }
    \end{algorithm}}


\section{Algorithms to be Considered}
    In this section we discuss the agents we want to evaluate for the task of algorithm control.
    We first discuss how to apply standard black-box optimization to the task of algorithm control.
    We then present agents that are capable of taking state information into account.
    
    \subsection{Black-Box Optimizer}
    In a standard black-box optimization setting, the optimizer interacts with an intended target
    by setting the configuration of the target at the beginning and waiting until the
    target returns the final reward signal. This is, e.g., the case in algorithm configuration. 
    The same setup can be easily extended to search for
    sequences of configurations for online configuration of the target. Instead of setting a
    \param once, the optimizer would have to set a sequence of \param values, once per time-step
    at which the target should switch its configuration.
    For sequences with $T$ such change points and large $T$, this drastically increases the configuration space, since 
    the optimizer would
    need to treat each individual parameter as $T$ different \param[s].
    In addition, black-box optimizers cannot observe the state information, which are required to learn instance-specific policies.
    
    \subsection{Context-oblivious Agents}
    \label{sub:coa}
    As a proof-of-concept
    \citeN{adriaensen-ijcai16} introduce context-oblivious agents
    that can take state information into account when selecting which action to play next.
    In their experiments the only state information they took into account
    was the history of the actions.
    
    To move their proposed agents from a black-box setting towards a white-box setting,
    during training the agents keep track of the number of times an action lead
    from one state to another, as well as the average reward this transition produced.
    This tabular approach limits the agents to small state and action spaces.
    The proposed agents include:
    \begin{itemize}
        \item URS: Selects an action uniformly at random.
        \item PURS: Selects a previously not selected action uniformly at random.
            Otherwise, actions are selected in proportion to the expected number of remaining steps.
        \item GR: Selects an action greedily based on the expected future reward.
    \end{itemize}
    During the evaluation phase, all agents greedily select the best action given
    the observations recorded during training.
    
    URS and GR both are equivalent to the two extremes of $\epsilon$-greedy
    Q-learning~\cite{watkins-ml92}, with $\epsilon=1$ and $\epsilon=0$ respectively.
    PURS leverages information about the expected trajectory length, but it does not include the
    observed reward signal in the decision making process. For tasks where every execution path
    has the same length (e.g. \emph{Counting}, \emph{Luby}, \emph{Sigmoid} and \emph{SigmoidMVA}),
    PURS would fail to produce a policy other than a uniform random one. 
    Further, when using PURS, we need to
    have some prior knowledge if shorter or longer trajectories should be preferred. For example
    on benchmarks like \emph{Fuzzy}, PURS is only able to find a meaningful
    policy if we know that longer sequences produce better rewards.
    
    \subsection{Reinforcement Learning}
    Reinforcement learning (RL) is a promising candidate to learn algorithm control policies in a data driven
    fashion because we can formulate algorithm control as an MDP
    and we can sample a large number of episodes given enough compute resources.
    An RL agent repeatedly interacts with the target algorithm by choosing some configurations at a given
    time-step and observing the state transition as well as the reward. Then, the RL agent updates its believe state about how the target algorithm will
    behave when using the chosen configuration at that time-step.
    Through these interactions, over
    time, the agent can typically find a policy that yields higher rewards.
    For small action and state spaces, RL agents can be easily implemented using
    table lookups, whereas for larger
    spaces, function approximation methods can make learning feasible.
    We evaluated $\epsilon$-greedy Q-learning~\cite{watkins-ml92} in the tabular setting
    as well as DQN using function approximations~\cite{mnih-icml16a}.

\section{Experimental Study}
    To compare black-box optimizers, context-oblivious agents~\cite{adriaensen-ijcai16} and
    reinforcement learning agents for algorithm control, we evaluated various agents
    on the benchmarks discussed above.

    	\begin{figure*}[tbp]
    	    \centering
    	        \subfloat[Counting][Counting]{\includegraphics[width=.33\textwidth]{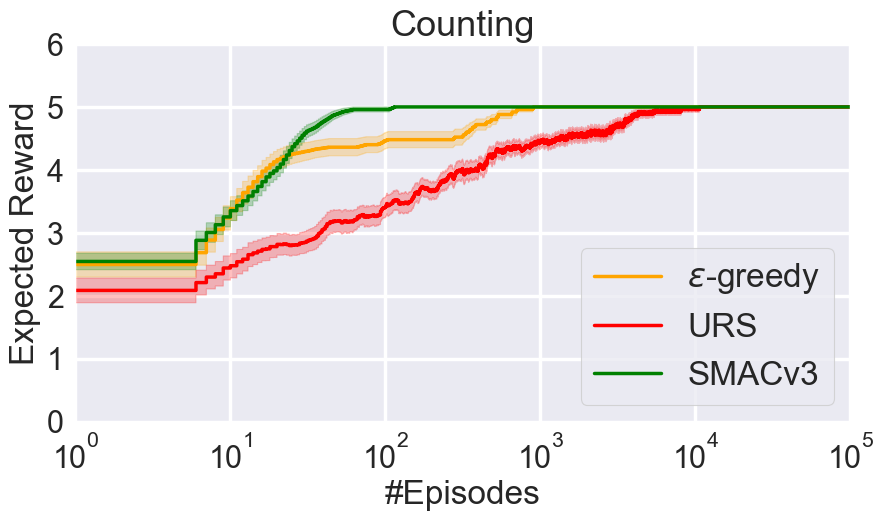}\label{fig:count}}
    	        \subfloat[Fuzzy][Fuzzy]{\includegraphics[width=.33\textwidth]{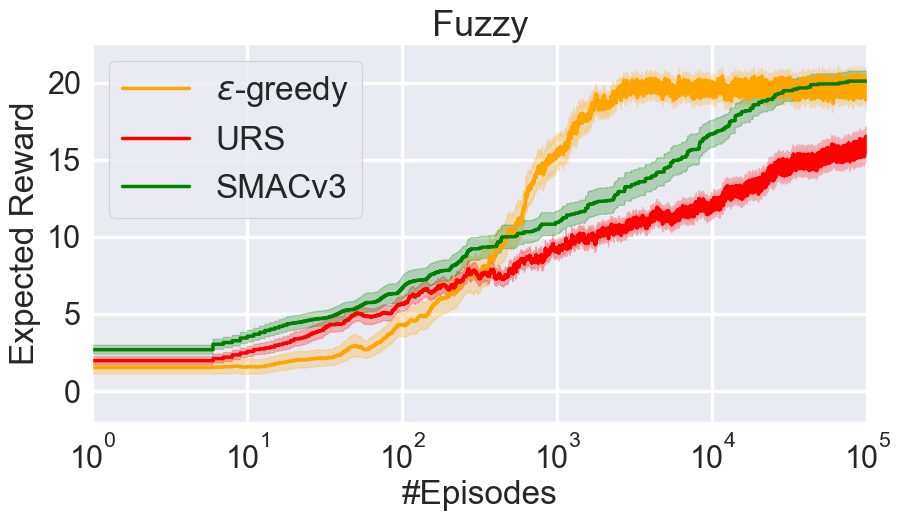}\label{fig:fuzzy}}
    	        \subfloat[Luby][Luby]{\includegraphics[width=.33\textwidth]{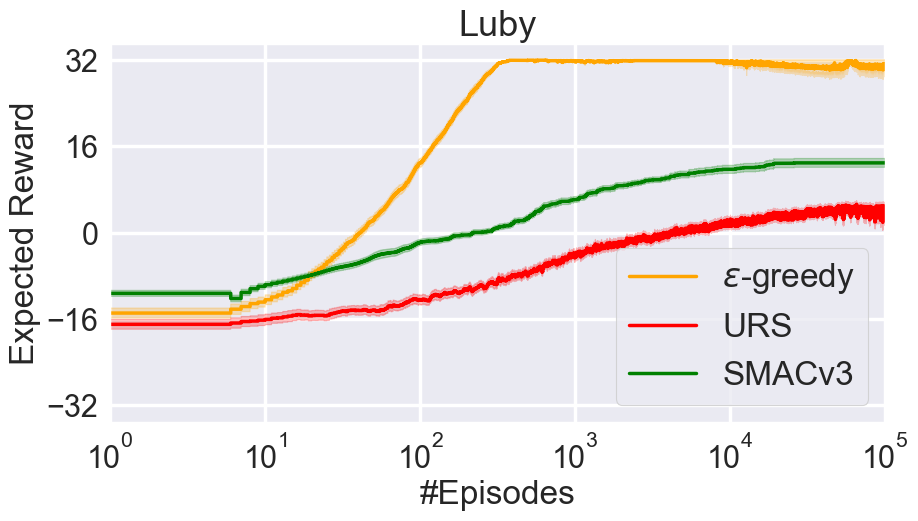}\label{fig:luby}}
    	    \caption{
    	    Results of SMAC, URS and tabular $\epsilon$-greedy Q-learning,
    	    on a set
    	    of discrete benchmarks. The $x$-axis depicts the number of episodes seen during training
    	    and the $y$-axis the gained reward. The lines depict the gained
    	    reward for each agent when evaluating it after the given number of training episodes with
    	    the solid line representing the mean reward over 25 repetitions and the shaded area the
    	    standard error. The presented lines are smoothed over a window of size 10. The results in
    	    \protect\subref{fig:count} are obtained on \emph{Counting}. The
    	    results in \protect\subref{fig:fuzzy} stem from \emph{Fuzzy} and
    	    the results in \protect\subref{fig:luby} depict the results on \emph{Luby}.
    	    }
    	    \label{fig:res_homogeneous_bench}
    	\end{figure*}
    \subsection{Setup}
	    We used SMAC~\cite{hutter-lion11a} in version 3 (SMACv3~\cite{smac-2017}) as a state-of-the-art algorithm configurator and black-box optimizer.
	    We implemented URS using simple tabular $1$-greedy Q-learning. We decided against using
	    PURS as it would only be applicable to \emph{Fuzzy}, see Section~\ref{sub:coa}.
	    
	    Q-learning based approaches were
	    evaluated using a discounting factor of $0.99$. On benchmarks with stochastic reward we set
	    the learning rate to $0.1$ and to $1.0$ otherwise. The $\epsilon$-greedy agent was trained
	    using a constant $\epsilon=0.1$.

        As \emph{Sigmoid} has continuous state features we include Q-learning using function
	    approximation in the form of a DQN~\cite{mnih-arxiv13} implemented in RLlib~\cite{liang-icml18}. 
	    We used the default configuration of the DQN in RLlib (0.6.6), i.e., 
	    a double dueling DQN where the target network is updated every 5 episodes and
	    the exploration fraction $\epsilon$ of the DQN is linearly decreased from $1.0$ to $0.02$.
	    We only changed the number of hidden units to 50 
	    and the training batch size and the timesteps
	    per training iteration to the episode length such that in each training iteration only one
	    episode is observed.
	    
	    In each training iteration each agent observed a full
	    episode. Training runs for all methods were repeated 25 times using different random seeds and
        each agent was evaluated after updating its policy. When evaluating on the deterministic benchmarks (\emph{Counting} and \emph{Luby}) only one evaluation run was performed. On
        the other benchmarks we performed $10$ evaluation runs of which we report the mean reward.
        When using a fixed instance set of size $100$ on \emph{Sigmoid} and \emph{SigmoidMVA}, we evaluated the
        agents once on each instance.
        
	    All experiments were run on a compute cluster with nodes equipped with two Intel Xeon E5-2630v4
        and 128GB memory running CentOS 7. The results on the
        benchmarks that do not model problem instances
        (\emph{Counting}, \emph{Fuzzy} and \emph{Luby}) are plotted in Figure~\ref{fig:res_homogeneous_bench}. The results for benchmarks with \task[s] (\emph{Sigmoid}
        and \emph{SigmoidMVA}) are shown in
        Figures~\ref{fig:res_heterogeneous_bench} and~\ref{fig:res_sigmva}.

        \subsection{Results}
        \paragraph{Counting}    	
    	Figure~\ref{fig:count} shows the evaluation results of SMAC,
    	URS, as well as
    	an $\epsilon$-greedy agent on \emph{Counting}. The agents are tasked with learning a policy
    	of length $T=5$ with $a_t = \left\{0, 1, 2, 3, 4\right\}$ for all $t \leq T$. 
    	On this simple benchmark, SMAC outperforms both other methods and learns the optimal policy after observing approximately $100$ episodes. This is in contrast to \citeN{adriaensen-ijcai16}, where
    	on this benchmark they evaluated black-box optimization for a static policy producing constant reward $1$.
    	The $\epsilon$-greedy
    	agent quickly learns policies in which $4$ out of $5$ choices are set correctly
    	but requires to observe approximately $900$ episodes until it learns the optimal
    	policy. URS purely exploratory behaviour prohibits quick learning of simple
    	policies, requiring close to $10^4$ episodes until it recovers the optimal policy.
    	
    	\paragraph{Fuzzy}  
    	The results of the agents' behaviours on \emph{Fuzzy} with $T=20$ are presented in
    	Figure~\ref{fig:fuzzy} and extend the findings of \citeN{adriaensen-ijcai16}. In such a noisy setting, $\epsilon$-greedy Q-learning is faster than SMAC in learning the
    	optimal policy, approaching it after roughly $2\,000$ episodes. However, SMAC is still able to learn the optimal policy after more than $24\,000$ episodes. URS has still not learned the optimal policy after $10^5$ episodes, only learning a policy
    	that chooses action $1$ approximately $16$ times before choosing action $0$.
    	
    	\paragraph{Luby}
    	Learning the optimal policy for \emph{Luby} requires the agent to learn a
    	policy of length $T=32$ with $a_t = \left\{0, 1, 2, 3, 4, 5\right\}$ for all $t\leq T$. 
    	The $\epsilon$-greedy agent already learns the optimal policy after
    	observing about $200$ episodes. In roughly the same amount of episodes SMAC found a
    	policy in which half of the choices are set correctly, and after observing $10^5$ episodes
    	it is able to find a policy that selects roughly $70\%$ of the actions correctly.
    	URS was roughly $10$ times slower in
    	learning a policy that achieves a reward of $0$, selecting half of the actions correctly after roughly $2\cdot10^3$ episodes (see Figure~\ref{fig:luby}); it found its final performance 100 times slower than SMAC.
    	
    	\begin{figure*}[tbp]
    	    \centering
    	        \subfloat[Sigmoid][Sigmoid]{\includegraphics[width=.33\textwidth]{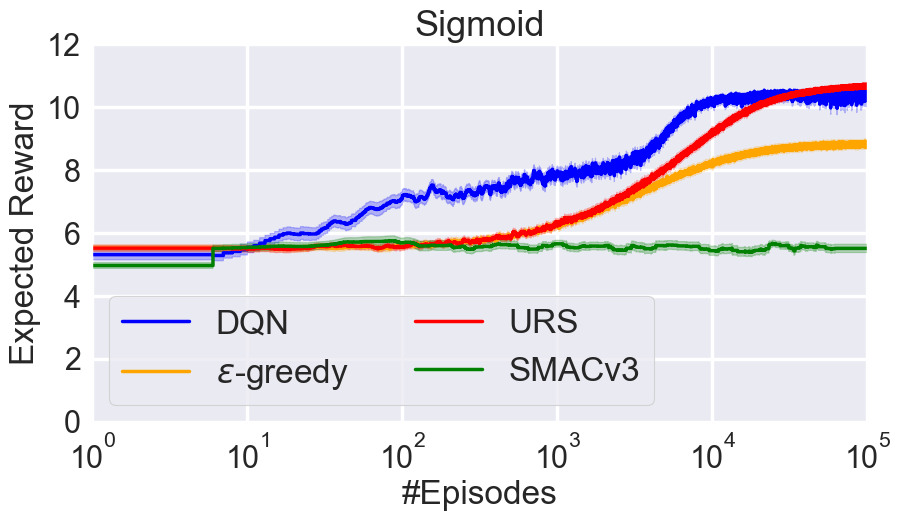}\label{fig:sig}}%
    	        \subfloat[Sigmoid Traing][Sigmoid Traing]{\includegraphics[width=.33\textwidth]{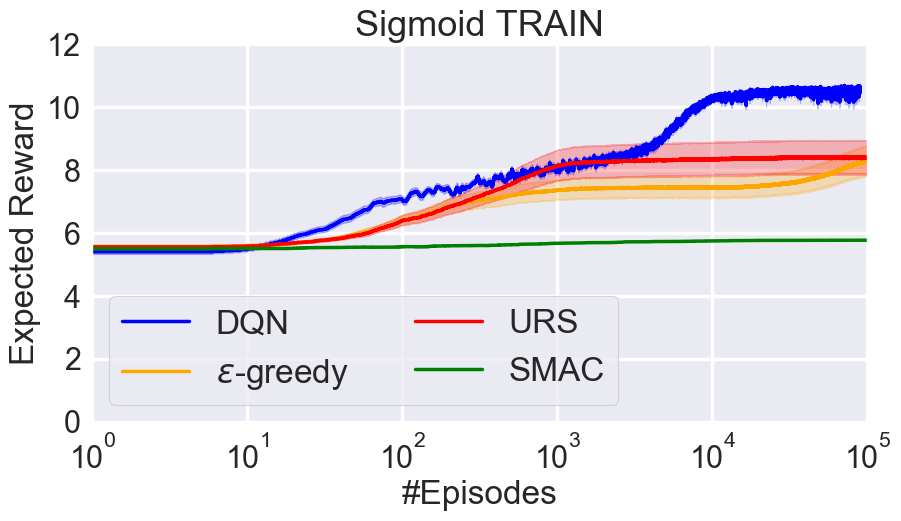}\label{fig:siginst}}%
    	        \subfloat[Sigmoid Test][Sigmoid Test]{\includegraphics[width=.33\textwidth]{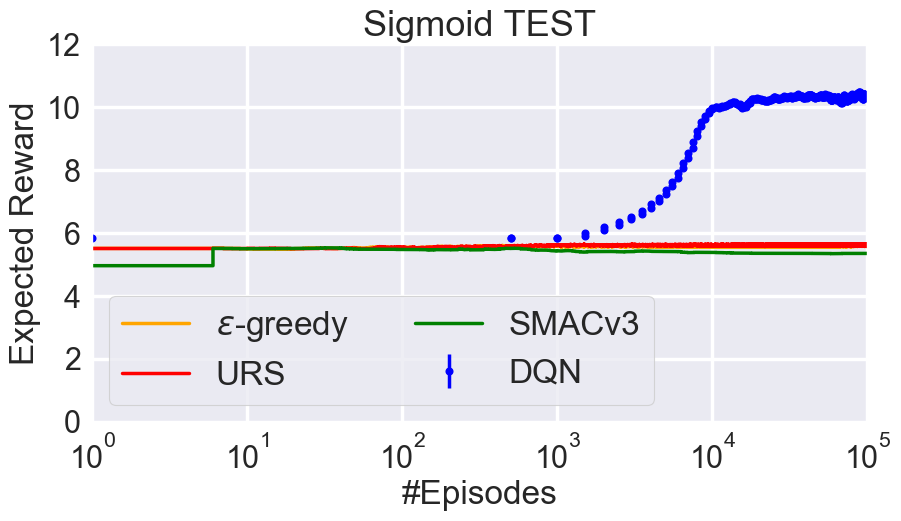}\label{fig:siginsttest}}
    	    \caption{
    	    Comparison of SMAC, URS and tabular $\epsilon$-greedy Q-learning,
    	    and a
    	    DQN on \emph{Sigmoid}. The $x$-axis depicts the number of episodes seen during training
    	    and the $y$-axis the gained reward. The lines depict the gained reward for each agent when
    	    evaluating it after the given number of training episodes with the solid line
    	    representing the mean reward over 25 training repetitions and the shaded area the standard
    	    error.
    	    To estimate the performance over the distribution of instances in
    	    \protect\subref{fig:sig_A}, we sample 10 random sigmoid functions when evaluating the
    	    agents. In the case of \protect\subref{fig:sig_B} we evaluate the agents on all $100$
    	    training instances.
    	    The presented lines are smoothed over a window of size 10. 
    	    \protect\subref{fig:sig} depicts results over a distribution of instances, 
    	    \protect\subref{fig:siginst} depicts results over a fixed set of training instances and
    	    \protect\subref{fig:siginsttest} the results on prior unseen test instances,
    	    evaluated every $500$ training episodes. For the tabular approaches the sate values
    	    have been rounded to the closest integer.
    	    }
    	    \label{fig:res_heterogeneous_bench}
    	\end{figure*}
    	
    	\paragraph{Sigmoid}
    	Results 
    	considering instances are shown in
    	Figure~\ref{fig:res_heterogeneous_bench}. In this setting the agents have to learn to adapt
    	their policies to the presented \task, with each policy of length $T=11$ and
    	$a_t = \left\{0, 1\right\}\forall t\leq T$.
    	For each episode an instance
    	can be either directly sampled or taken out of a set of instances stemming from
    	the same distribution. Therefore an agent that learns
    	policies dependent on the task can achieve a maximal reward of $11$ and black-box optimizers can only achieve at most $6$.
    	To allow the tabular Q-learning
    	approaches to work on this continuous state-space we round the scaling factor and inflection
    	point to the closest integer values.
    	
    	Figure~\ref{fig:sig} shows the gained reward of the agents when randomly drawing new instances
    	in each training iteration. We can observe that all agents received a reward of
    	roughly $5.5$ for
    	randomly selecting which actions to play. The DQN quickly began to learn faster than either
    	of the tabular approaches, receiving a reward of $~8$ before $\epsilon$-greedy and
    	URS begin to learn an improving policy. After roughly $10^4$ training episodes the DQN
    	learns policies that adapt to the instance at hand, whereas the $\epsilon$-greedy agent
    	gets stuck in a local optimum. Due to being completely exploratory, URS does not exhibit the
    	same behaviour and can continue to improve its policy before learning the optimal policy
    	after roughly $2\cdot10^4$ training episodes.
    	SMAC is unable to find a policy that is able to adapt to the instances at
    	hand. This is due to the optimizer not being able to distinguish between a positive and
    	negative slope of the sigmoid. Therefore, it cannot decide if it should start
    	a policy with action $0$ or $1$ before switching to the other. Furthermore, the agent does not
    	know the inflection point and can only guess when to switch from one action to the other.
    	
    	It is most often not the case that we have an entire distribution of instances at our disposal, but
    	only a finite set of instances sampled from an unknown distribution.
    	To include this setting in our evaluation, we sampled $100$ training and $100$ test instances
    	from the same distribution used before. The reported results here give the performance
    	across this whole training or test set. On the training set, the results for the DQN as well as SMAC look very similar
    	to the results for the distribution of instances.
    	Both tabular agents learn much
    	faster since the possible state-space is much smaller. However both agents get stuck in a local optimum and are unable to recover the optimal policy.
    	
    	On the test instances, the tabular agents are incapable of generalization (see Figure~\ref{fig:siginsttest}), but, using function approximation, DQN is able
    	to generalize. At first, DQN overfits on a few training instances, before it learns a robust policy for many training instances that generalizes to the test instances.
    	
    	\begin{figure}[tbp]
    	    \centering
    	    \includegraphics[width=.4\textwidth]{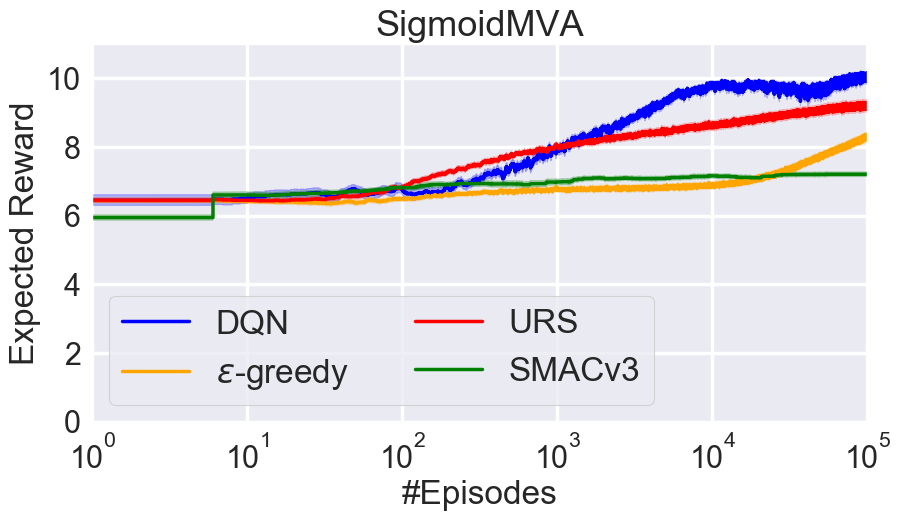}
    	    \caption{Comparison of the agents on \emph{SigmoidMVA}.
    	    The $x$ and $y$-axis show the number of episodes and the gained reward respectively. The lines depict the reward for each agent when
    	    evaluating it after the given number of training episodes where the line
    	    represents the mean reward over 25 training repetitions and the shaded area the standard
    	    error, smoothed over a window of size 10.}
    	    \label{fig:res_sigmva}
	    \end{figure}

	    \paragraph{SigmoidMVA}
	     The results on \emph{SigmoidMVA} are shown in Figure~\ref{fig:res_sigmva}.
	     Similar to \emph{Sigmoid}, agents need to adapt their
	     policy to a sampled instance, however, on an extended action space of size $5$.
	     Again URS benefits from its random sampling behavior, whereas the $\epsilon$-greedy agent needs to observe roughly $10^4$ episodes before
	     improving over a random policy. Without any state information SMAC struggles
	     to find a meaningful policy and the DQN is capable of adjusting the policy
	     to the instance at hand even on this higher-dimensional space.
	
\section{Conclusion}
	To the best of our knowledge we are the first to formalize the algorithm control problem as
	a contextual MDP, explicitly taking problem \task[s] into account. 
	To study different agents types for the problem of algorithm control with instances,
	we present new white-box benchmarks.
	Using these benchmarks, we showed that black-box
	optimization is a feasible candidate to learn policies for simple action spaces.
	With increasing complexity of the optimal policy however,  black-box optimizers struggle to
	learn such an optimal policy. In contrast, reinforcement learning is a suitable candidate for
	learning more complex sequences. 
	If heterogeneous instances are considered,  black-box optimizers
	might struggle to learn any policy that is better than a random policy.
	In contrast, RL agents
	making use of state information are able to adapt their policies to the problem
	instance, which demonstrates the potential of applying RL to algorithm control. 
	
	The presented white-box benchmarks are a first step towards scenarios resembling real algorithm control for hard-combinatorial problem solvers on a set of instances. In future work,
	we plan to extend our benchmarks considering mixed spaces of categorical and continuous hyperparameters
	and conditional dependencies. Furthermore, we plan to train cheap-to-evaluate surrogate benchmarks based on data gathered from real
	algorithm runs~\cite{eggensperger-mlj18a}.

\paragraph*{Acknowledgments}
The authors acknowledge funding by the Robert Bosch GmbH, support by the state of Baden-Württemberg through bwHPC and the German Research Foundation (DFG) through grant no. INST 39/963-1 FUGG.

\bibliographystyle{named}
\bibliography{arxive_submission}

\begin{thebibliography}{}

\bibitem[\protect\citeauthoryear{Adriaensen and
  Now{\'e}}{2016}]{adriaensen-ijcai16}
S.~Adriaensen and A.~Now{\'e}.
\newblock Towards a white box approach to automated algorithm design.
\newblock In {\em Proc. of IJCAI'16}, pages 554--560, 2016.

\bibitem[\protect\citeauthoryear{Andrychowicz \bgroup \em et al.\egroup
  }{2016}]{andrychowicz-neurips16}
M.~Andrychowicz, M.~Denil, S.~Gomez, M.~W. Hoffman, D.~Pfau, T.~Schaul,
  B.~Shillingford, and N.~De Freitas.
\newblock Learning to learn by gradient descent by gradient descent.
\newblock In {\em Proc. of NIPS'16}, pages 3981--3989, 2016.

\bibitem[\protect\citeauthoryear{Ans{\'o}tegui \bgroup \em et al.\egroup
  }{2009}]{ansotegui-cp09a}
C.~Ans{\'o}tegui, M.~Sellmann, and K.~Tierney.
\newblock A gender-based genetic algorithm for the automatic configuration of
  algorithms.
\newblock In {\em Proc. of CP'09}, pages 142--157, 2009.

\bibitem[\protect\citeauthoryear{Ans{\'o}tegui \bgroup \em et al.\egroup
  }{2017}]{ansotegui-aaai17}
C.~Ans{\'o}tegui, J.~Pon, M.~Sellmann, and K.~Tierney.
\newblock Reactive dialectic search portfolios for maxsat.
\newblock In {\em Proc. of AAAI'17}, 2017.

\bibitem[\protect\citeauthoryear{Battiti and Campigotto}{2011}]{battiti-11}
R.~Battiti and P.~Campigotto.
\newblock An investigation of reinforcement learning for reactive search
  optimization.
\newblock In {\em Autonomous Search}, pages 131--160. Springer, 2011.

\bibitem[\protect\citeauthoryear{Battiti \bgroup \em et al.\egroup
  }{2008}]{battiti-book08}
Roberto Battiti, Mauro Brunato, and Franco Mascia.
\newblock {\em Reactive search and intelligent optimization}, volume~45.
\newblock Springer Science \& Business Media, 2008.

\bibitem[\protect\citeauthoryear{Chen \bgroup \em et al.\egroup
  }{2017}]{chen-arxiv16}
Y.~Chen, M.~W. Hoffman, S.~G. Colmenarejo, M.~Denil, T.~P. Lillicrap,
  M.~Botvinick, and N.~De Freitas.
\newblock Learning to learn without gradient descent by gradient descent.
\newblock In {\em Proc. of ICML'17}, pages 748--756, 2017.

\bibitem[\protect\citeauthoryear{Daniel \bgroup \em et al.\egroup
  }{2016}]{daniel-aaai16}
C.~Daniel, J.~Taylor, and S.~Nowozin.
\newblock Learning step size controllers for robust neural network training.
\newblock In {\em Proc. of AAAI'16}, 2016.

\bibitem[\protect\citeauthoryear{Doerr and Doerr}{2018}]{doerr-arxi18}
B.~Doerr and C.~Doerr.
\newblock Theory of parameter control for discrete black-box optimization:
  Provable performance gains through dynamic parameter choices.
\newblock arXiv:1804.05650, 2018.

\bibitem[\protect\citeauthoryear{Eggensperger \bgroup \em et al.\egroup
  }{2018}]{eggensperger-mlj18a}
K.~Eggensperger, M.~Lindauer, H.~H. Hoos, F.~Hutter, and K.~Leyton{-}Brown.
\newblock Efficient benchmarking of algorithm configurators via model-based
  surrogates.
\newblock {\em Machine Learning}, 107(1):15--41, 2018.

\bibitem[\protect\citeauthoryear{Fawcett \bgroup \em et al.\egroup
  }{2011}]{fawcett-icasp11a}
C.~Fawcett, M.~Helmert, H.~Hoos, E.~Karpas, G.~Roger, and J.~Seipp.
\newblock Fd-autotune: Domain-specific configuration using fast-downward.
\newblock In {\em Proc. of ICAPS'11}, 2011.

\bibitem[\protect\citeauthoryear{Hutter \bgroup \em et al.\egroup
  }{2010}]{hutter-cpaior10a}
F.~Hutter, H.~Hoos, and K.~Leyton-Brown.
\newblock Automated configuration of mixed integer programming solvers.
\newblock In {\em Proc. of CPAIOR'10}, pages 186--202, 2010.

\bibitem[\protect\citeauthoryear{Hutter \bgroup \em et al.\egroup
  }{2011}]{hutter-lion11a}
F.~Hutter, H.~Hoos, and K.~Leyton-Brown.
\newblock Sequential model-based optimization for general algorithm
  configuration.
\newblock In {\em Proc. of LION'11}, pages 507--523, 2011.

\bibitem[\protect\citeauthoryear{Hutter \bgroup \em et al.\egroup
  }{2017}]{hutter-aij17a}
F.~Hutter, M.~Lindauer, A.~Balint, S.~Bayless, H.~Hoos, and K.~Leyton-Brown.
\newblock The configurable {SAT} solver challenge ({CSSC}).
\newblock 243:1--25, 2017.

\bibitem[\protect\citeauthoryear{Kadioglu \bgroup \em et al.\egroup
  }{2010}]{kadioglu-ecai10}
S.~Kadioglu, Y.~Malitsky, M.~Sellmann, and K.~Tierney.
\newblock {ISAC} - instance-specific algorithm configuration.
\newblock In {\em Proc. of ECAI'10}, pages 751--756, 2010.

\bibitem[\protect\citeauthoryear{Karafotias \bgroup \em et al.\egroup
  }{2015}]{karafotias-ec15}
G.~Karafotias, M.~Hoogendoorn, and A.~E. Eiben.
\newblock Parameter control in evolutionary algorithms: Trends and challenges.
\newblock {\em IEEE Transactions on Evolutionary Computation}, 19(2):167--187,
  2015.

\bibitem[\protect\citeauthoryear{Kingma and Welling}{2014}]{Kingma-iclr14}
D.~Kingma and M.~Welling.
\newblock Auto-encoding variational bayes.
\newblock In {\em Proc. of ICLR'14}, 2014.

\bibitem[\protect\citeauthoryear{Leyton-Brown \bgroup \em et al.\egroup
  }{2009}]{leyton-brown-acm09a}
K.~Leyton-Brown, E.~Nudelman, and Y.~Shoham.
\newblock Empirical hardness models: Methodology and a case study on
  combinatorial auctions.
\newblock {\em Journal of the ACM}, 56(4):1--52, 2009.

\bibitem[\protect\citeauthoryear{Li and Malik}{2017}]{li-arxiv16}
K.~Li and J.~Malik.
\newblock Learning to optimize.
\newblock In {\em Proc. of ICLR'17}, 2017.

\bibitem[\protect\citeauthoryear{Liang \bgroup \em et al.\egroup
  }{2018}]{liang-icml18}
E.~Liang, R.~Liaw, R.~Nishihara, P.~Moritz, R.~Fox, K.~Goldberg, J.~Gonzalez,
  M.~Jordan, and I.~Stoica.
\newblock Rllib: Abstractions for distributed reinforcement learning.
\newblock In {\em Proc. of ICML'18}, pages 3059--3068, 2018.

\bibitem[\protect\citeauthoryear{Lindauer \bgroup \em et al.\egroup
  }{2017}]{smac-2017}
M.~Lindauer, K.~Eggensperger, M.~Feurer, S.~Falkner, A.~Biedenkapp, and
  F.~Hutter.
\newblock Smac v3: Algorithm configuration in python.
\newblock \url{https://github.com/automl/SMAC3}, 2017.

\bibitem[\protect\citeauthoryear{L{\'{o}}pez{-}Ib{\'{a}}{\~{n}}ez \bgroup \em
  et al.\egroup }{2016}]{lopez-ibanez-orp16}
M.~L{\'{o}}pez{-}Ib{\'{a}}{\~{n}}ez, J.~Dubois-Lacoste, L.~Perez Caceres,
  M.~Birattari, and T.~St{\"{u}}tzle.
\newblock The irace package: Iterated racing for automatic algorithm
  configuration.
\newblock {\em Operations Research Perspectives}, 3:43--58, 2016.

\bibitem[\protect\citeauthoryear{Loshchilov and
  Hutter}{2017}]{loshchilov-iclr17a}
I.~Loshchilov and F.~Hutter.
\newblock Sgdr: Stochastic gradient descent with warm restarts.
\newblock In {\em Proc. of ICLR'17}, 2017.

\bibitem[\protect\citeauthoryear{Luby \bgroup \em et al.\egroup
  }{1993}]{luby-ipl93}
M.~Luby, A.~Sinclair, and D.~Zuckerman.
\newblock Optimal speedup of las vegas algorithms.
\newblock {\em Information Processing Letters}, 47(4):173--180, 1993.

\bibitem[\protect\citeauthoryear{Mnih \bgroup \em et al.\egroup
  }{2013}]{mnih-arxiv13}
V.~Mnih, K.~Kavukcuoglu, D.~Silver, A.~Graves, I.~Antonoglou, D.~Wierstra, and
  M.~Riedmiller.
\newblock Playing atari with deep reinforcement learning.
\newblock arXiv:1312.5602, 2013.

\bibitem[\protect\citeauthoryear{Mnih \bgroup \em et al.\egroup
  }{2016}]{mnih-icml16a}
V.~Mnih, A.~Badia, M.~Mirza, A.~Graves, T.~Lillicrap, T.~Harley, D.~Silver, and
  K.~Kavukcuoglu.
\newblock Asynchronous methods for deep reinforcement learning.
\newblock In {\em Proc. of ICML'16}, pages 1928--1937, 2016.

\bibitem[\protect\citeauthoryear{Moulines and Bach}{2011}]{moulines-neurips11}
E.~Moulines and F.~R. Bach.
\newblock Non-asymptotic analysis of stochastic approximation algorithms for
  machine learning.
\newblock In {\em Proc. of NIPS'11}, pages 451--459, 2011.

\bibitem[\protect\citeauthoryear{Rice}{1976}]{rice76a}
J.~Rice.
\newblock The algorithm selection problem.
\newblock {\em Advances in Computers}, 15:65--118, 1976.

\bibitem[\protect\citeauthoryear{Sakurai \bgroup \em et al.\egroup
  }{2010}]{sakurai-sitis2010}
Y.~Sakurai, K.~Takada, T.~Kawabe, and S.~Tsuruta.
\newblock A method to control parameters of evolutionary algorithms by using
  reinforcement learning.
\newblock In {\em Proc. of {SITIS}}, pages 74--79, 2010.

\bibitem[\protect\citeauthoryear{Schaul \bgroup \em et al.\egroup
  }{2013}]{schaul-icml13}
T.~Schaul, S.~Zhang, and Y.~LeCun.
\newblock {No More Pesky Learning Rates}.
\newblock In {\em Proc. of ICML'13}, 2013.

\bibitem[\protect\citeauthoryear{Schneider and Hoos}{2012}]{hoos-lion12a}
M.~Schneider and H.~Hoos.
\newblock Quantifying homogeneity of instance sets for algorithm configuration.
\newblock In {\em Proc. of LION'12}, pages 190--204, 2012.

\bibitem[\protect\citeauthoryear{Singh \bgroup \em et al.\egroup
  }{2015}]{singh-ieee15}
B.~Singh, S.~De, Y.~Zhang, T.~Goldstein, and G.~Taylor.
\newblock Layer-specific adaptive learning rates for deep networks.
\newblock In {\em Proc. of ICMLA'15}, pages 364--368, 2015.

\bibitem[\protect\citeauthoryear{Snoek \bgroup \em et al.\egroup
  }{2012}]{snoek-nips12a}
J.~Snoek, H.~Larochelle, and R.~Adams.
\newblock Practical {B}ayesian optimization of machine learning algorithms.
\newblock In {\em Proc. of NIPS'12}, pages 2960--2968, 2012.

\bibitem[\protect\citeauthoryear{Watkins and Dayan}{1992}]{watkins-ml92}
C.~Watkins and P.~Dayan.
\newblock Q-learning.
\newblock {\em Machine learning}, 8(3-4):279--292, 1992.

\bibitem[\protect\citeauthoryear{Xu \bgroup \em et al.\egroup
  }{2010}]{xu-aaai10a}
L.~Xu, H.~Hoos, and K.~Leyton-Brown.
\newblock Hydra: Automatically configuring algorithms for portfolio-based
  selection.
\newblock In {\em Proc. of AAAI'10}, pages 210--216, 2010.

\end{thebibliography}
\end{document}